\documentclass{article}
\usepackage[T1]{fontenc}
\usepackage{fix-cm}
\usepackage{iclr2027_conference,times}

\usepackage{amsmath,amsfonts,bm}

\def\eqref#1{equation~\ref{#1}}

\def\1{\bm{1}}

\DeclareMathAlphabet{\mathsfit}{\encodingdefault}{\sfdefault}{m}{sl}
\SetMathAlphabet{\mathsfit}{bold}{\encodingdefault}{\sfdefault}{bx}{n}

\usepackage{amsmath}
\usepackage{amssymb}
\usepackage{booktabs}
\usepackage{colortbl}
\usepackage{float}
\usepackage{graphicx}
\usepackage{wrapfig}
\usepackage{multirow}
\usepackage{needspace}
\usepackage{caption}
\usepackage[hyphens]{url}
\usepackage{xurl}
\usepackage{hyperref}
\hypersetup{colorlinks=true,linkcolor=black,citecolor=black,urlcolor=blue}

\newcommand{\method}{\textsc{BLPO}}

\title{\method: Beyond-Local Policy Optimization for Long-Horizon LLM Agents}

\author{%
Xuanren Chen$^{1}$\quad Chongyang Tao$^{1,*}$\quad Wenpeng Hu\quad Yong Wang$^{2}$\\[2pt]
Bo Deng$^{2}$\quad Xinlei Zheng$^{1}$\quad Shuai Ma$^{1}$\\[5pt]
{\small $^{1}$Beihang University\quad $^{2}$Alibaba Group}\\[4pt]
{\small\texttt{\{cxrbuaa,chongyang,sy2406103,xinleizh,mashuai\}@buaa.edu.cn}}\\
{\small\texttt{wenpeng.hu@pku.edu.cn}\quad\texttt{wangyong.lz@alibabainc.com}}\\[3pt]
{\small $^{*}$Corresponding author.}}

\iclrfinalcopy
\hypersetup{pdftitle={BLPO: Beyond-Local Policy Optimization for Long-Horizon LLM Agents},pdfauthor={Xuanren Chen, Chongyang Tao, Wenpeng Hu, Yong Wang, Bo Deng, Xinlei Zheng, Shuai Ma}}

\usepackage{mdframed}
\makeatletter
\renewcommand{\@maketitle}{%
  \begingroup
  \centering
  {\LARGE\bfseries\@title\par}
  \vspace{10pt}
  {\normalsize\normalfont\begin{tabular}{@{}c@{}}\@author\end{tabular}\par}
  \vspace{10pt}
  \endgroup
}
\makeatother

\begin{document}

\maketitle
\lhead{Preprint}

\begin{abstract}
Long-horizon tasks require large language model (LLM) agents to coordinate interdependent actions, a capability central to automating complex workflows. Stepwise group-based reinforcement learning provides finer credit than trajectory-level supervision, but reliable comparisons must account for differences in interaction history. Enforcing this consistency within the limited rollouts of each task can leave insufficient support for estimating action preferences. We propose Beyond-Local Policy Optimization (BLPO), which estimates task-progress advantages from experience shared across tasks and training updates. We summarize trajectory prefixes into task-progress states and abstract actions by their function, allowing corresponding decisions from different tasks to contribute to shared estimates. To control bias and noise from heterogeneous tasks, we accumulate context-relative returns in a global memory with balanced aggregation and reliability weighting. These global advantages complement local comparisons over normalized actions and trajectory-level credit without a learned critic or additional rollouts. Experiments on ALFWorld and WebShop with 1.5B and 7B policies show improved task success over the evaluated baselines under matched training budgets. Further analyses show faster learning and shorter trajectories with little computational overhead.
Code is available at \mbox{\href{https://github.com/Tdadada/BLPO}{\nolinkurl{github.com/Tdadada/BLPO}}}.
\end{abstract}

\section{Introduction}

Large language model (LLM) agents extend automated assistance from answering individual queries to executing tasks in external environments~\citep{yao2023reactsynergizingreasoningacting,zhou2024webarena}. In applications such as travel planning and computational research, they offer the prospect of delegating entire workflows that otherwise require substantial human coordination~\citep{shen2026tripbench,li2026longhorizonterminalbench}. Realizing this potential requires solving long-horizon tasks, where earlier actions establish prerequisites for later steps and environmental feedback informs subsequent decisions~\citep{zhu2026longact}. Agents must therefore coordinate their actions over extended interactions, making the ability to learn policies that sustain progress toward task completion central to their practical utility.

Recent efforts to improve agents' performance on long-horizon tasks have increasingly explored learning from interaction, particularly through reinforcement learning (RL)~\citep{zhou2024archer,wang2025ragen,jin2025searchr1,he2026agentlightning}. Among these approaches, Group Relative Policy Optimization (GRPO) compares trajectory returns to optimize the policy without a separate critic~\citep{shao2024deepseekmathpushinglimitsmathematical}. However, assigning one advantage to all actions can reinforce errors in successful trajectories and suppress useful steps in failed ones. When all rollouts receive the same failure reward, their relative advantages vanish entirely~\citep{yang2026progpo}. To distinguish individual contributions, GiGPO compares actions taken from matching observed states across rollouts~\citep{feng2025groupingrouppolicyoptimizationllm}. Such comparisons can nevertheless mix different interaction histories, motivating HGPO to construct groups with varying degrees of historical consistency and adaptively combine their advantages~\citep{he2026hierarchyofgroupspolicyoptimizationlonghorizon}.

Despite these refinements, requiring greater historical consistency can leave the limited rollouts of a task insufficient for reliable comparisons~\citep{he2026hierarchyofgroupspolicyoptimizationlonghorizon}. ECPO improves the reliability of local action estimates~\citep{li2026ecpo}, while BiPACE expands comparison groups by aggregating similar states~\citep{wang2026bipace}; both still estimate credit within sampled rollout groups. Experience from related tasks offers additional support. Histories involving different entities may represent equivalent task progress, while different concrete actions may serve the same function. Grouping decisions on this basis allows comparisons to accumulate across tasks and training updates. Such sharing, however, must separate action effects from differences in task difficulty and prevent frequently visited contexts from dominating the estimates. Reliable global credit therefore requires correcting these biases while controlling how strongly uncertain comparisons influence policy updates.

Building on this perspective, we propose Beyond-Local Policy Optimization (BLPO), which estimates task-progress advantages by pooling experience across tasks and training updates. To construct these comparisons, predefined environment-specific rules summarize task progress from prior interactions and map concrete actions to their goal-relative functions. Applied automatically across task instances, these rules define comparison groups, while their action preferences are estimated from rollout returns. To control the heterogeneity introduced by sharing, we center returns within concrete contexts and balance their contributions to a decaying statistical memory. Preferences derived from this memory are weighted by their support and uncertainty, limiting the influence of unreliable estimates. The resulting task-progress advantages complement local comparisons over normalized actions, while trajectory advantages preserve supervision of overall task completion. All three signals enter a unified policy objective without a learned critic or additional rollouts. Experiments on ALFWorld and WebShop show improved task success with 1.5B and 7B policies, including a 2.48 percentage point gain over HGPO on WebShop with the 1.5B backbone. Further analyses support cross-task experience sharing and show faster learning with little computational overhead.

Our contributions are threefold:
\begin{itemize}
    \item We propose BLPO, a critic-free policy optimization method that reuses task-progress-aligned experience across tasks and training updates for step-level credit assignment, without additional rollouts.
    \item We develop a context-balanced global advantage estimator that reduces bias from context difficulty and repeated visits. Support and uncertainty scale the resulting credit to limit the influence of unreliable comparisons.
    \item Experiments on ALFWorld and WebShop show improved task success with both 1.5B and 7B policies. Controlled comparisons under the same abstraction support the benefits of cross-task sharing, while learning and cost analyses show faster ALFWorld learning with little additional computation.
\end{itemize}

\section{Related Work}

\paragraph{Reinforcement Learning for LLM Agents.}
Research on agentic RL increasingly treats sustained interaction as the basis for policy learning. ArCHer combines utterance-level value estimation with token-level optimization~\citep{zhou2024archer}, while RAGEN and Search-R1 develop training around multi-turn trajectories~\citep{wang2025ragen,jin2025searchr1}. As training extends to longer interactions, AgentGym-RL introduces a curriculum that progressively increases interaction length~\citep{xi2025agentgymrl}. Complementing these algorithmic developments, Agent Lightning v1.0 addresses the systems integration of RL with existing agents while preserving the execution harness's control of the interaction loop~\citep{he2026agentlightning}.

\paragraph{Credit Assignment in Agentic RL.}
Within these training frameworks, a central question is how task outcomes should supervise individual decisions. RLOO and GRPO offer relative estimates without a learned critic, but their direct trajectory-level application does not distinguish intermediate actions~\citep{ahmadian2024backbasicsrevisitingreinforce,shao2024deepseekmathpushinglimitsmathematical}. GiGPO makes these comparisons more granular by grouping decisions with matching observations~\citep{feng2025groupingrouppolicyoptimizationllm}. HGPO further conditions groups on interaction history, addressing inconsistencies hidden by observation matching~\citep{he2026hierarchyofgroupspolicyoptimizationlonghorizon}. Such finer comparisons also raise questions of statistical support and reliability. BiPACE increases support by grouping states in actor representation space~\citep{wang2026bipace}, whereas ECPO calibrates comparisons over equivalent actions using support and variability~\citep{li2026ecpo}. Alongside these group estimators, GraphGPO derives credit from goal distances in trajectory graphs~\citep{cheng2026trajectorylevelattributiongraphbasedcredit}, while HCAPO refines action values through hindsight reasoning~\citep{tan2026hindsightcreditassignmentlonghorizon}.

\paragraph{Experience Reuse for Agent Learning.}
The experience available for learning can also extend beyond current rollouts. One approach supplies earlier experience as policy context through retrieved step traces in RAPO and evolving skill documents in SkillRise~\citep{zhang2026rapo,yao2026skillrise}. Rather than exposing history through the policy input, TIGPO retains it in persistent transition graphs, combining past and current transitions for credit estimation~\citep{gan2026tigpo}. This form of temporal reuse remains confined to individual task instances. BLPO instead matches task progress and action function to pool comparisons across tasks and policy updates. To manage the resulting heterogeneity, we adjust returns within concrete contexts, balance their contributions, and attenuate uncertain comparisons.

\begin{figure}[t]
    \centering
    \includegraphics[width=\textwidth]{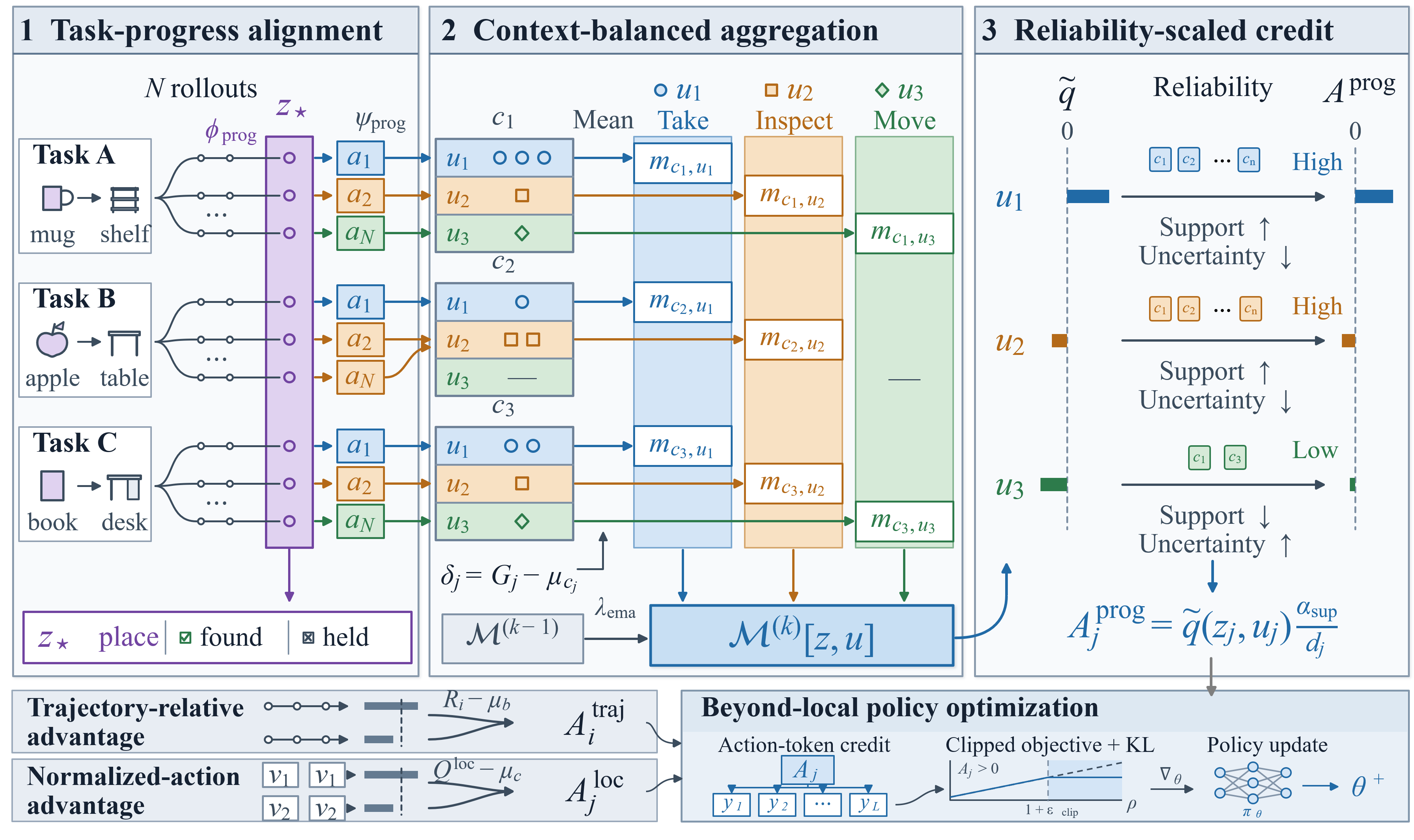}
    \caption{Overview of \method{}. (1) Trajectory histories and concrete actions from different tasks are mapped to shared task-progress states and action functions. (2) Within each concrete context, returns are centered and averaged per abstract state--action pair to update a global memory across training iterations. (3) Action contrasts estimated from this memory are scaled by effective support and uncertainty. The resulting task-progress advantages are combined with trajectory-relative advantages and local advantages over normalized actions to optimize the policy.}
    \label{fig:method-overview}
\end{figure}

\section{Preliminaries}
\label{sec:preliminaries}

For a task goal $g_i$, a trajectory $\tau_i$ consists of $T_i$ sequential decisions. At step $t$, the policy generates a textual response $a^{\mathrm{raw}}_{i,t}\sim\pi_\theta(\cdot\mid x_{i,t})$, where $x_{i,t}$ contains the goal, current observation $s_{i,t}$, and interaction history retained within the policy's context window. The response is parsed into an environment action $a_{i,t}$, whose execution yields feedback $e_{i,t}$, including the next observation, reward $r_{i,t}$, termination status, and format and admissibility information. The complete interaction prefix available before the current action is
\begin{equation}
    \mathcal{H}_{i,t}=
    \bigl(g_i,s_{i,1},a_{i,1},e_{i,1},\ldots,
    a_{i,t-1},e_{i,t-1},s_{i,t}\bigr).
    \label{eq:interaction-history}
\end{equation}
Unlike the policy input $x_{i,t}$, this prefix records all preceding interactions, including those outside the context window. We use $j=(i,t)$ to index an individual decision and write decision-specific quantities with either subscript.

\section{BLPO}
\label{sec:method}

As summarized in Figure~\ref{fig:method-overview}, we combine trajectory relative advantages (Sec.~\ref{sec:trajectory-advantage}) and local advantages over normalized actions (Sec.~\ref{sec:step-advantage}) with task-progress advantages from global experience (Sec.~\ref{sec:progress-advantage}). The global estimator pools context-adjusted returns across tasks and updates, accounting for support and uncertainty. Together, these signals define the policy objective in Section~\ref{sec:blpo-objective}.

\subsection{Trajectory Relative Advantage}
\label{sec:trajectory-advantage}

For each task instance, we sample $N$ trajectories from $\pi_{\theta_{\mathrm{old}}}$ under the same initial state. Let $b_i$ identify the rollout group containing $\tau_i$. Following GRPO~\citep{shao2024deepseekmathpushinglimitsmathematical}, we normalize trajectory returns within each group:
\begin{equation}
    R_i=\sum_{t=1}^{T_i}r_{i,t},
    \qquad
    A^{\mathrm{traj}}_i=
    \frac{R_i-\mu_{b_i}}{\sigma_{b_i}+\epsilon},
    \label{eq:trajectory-advantage}
\end{equation}
where $\mu_{b_i}$ and $\sigma_{b_i}$ are the group mean and standard deviation, and $\epsilon>0$ ensures numerical stability. Rewards follow the shared convention in Appendix~\ref{app:rollout-reward}, including the format penalty. The resulting advantage is shared by all decisions in the trajectory.

For the subsequent step comparisons, we discount future rewards to reduce the influence of feedback separated from the current decision by many intervening actions:
\begin{equation}
    G_j=\sum_{t'=t}^{T_i}\gamma^{t'-t}r_{i,t'},
    \qquad j=(i,t),\quad\gamma\in(0,1].
    \label{eq:step-return}
\end{equation}

\subsection{Step Advantage over Normalized Actions}
\label{sec:step-advantage}

To distinguish individual decisions, we compare actions under the same task and observation. However, advantages computed from individual returns can assign opposite signs to equivalent actions when subsequent decisions differ. We therefore pool their returns before estimating local credit.

For decision $j=(i,t)$, we define
\begin{equation}
    c_j=(b_i,s_j),
    \qquad
    v_j=\operatorname{NormAct}(a_j^{\mathrm{raw}},e_j).
    \label{eq:concrete-mapping}
\end{equation}
The context $c_j$ matches the rollout group and current observation without conditioning on earlier history. The action mapping removes textual variations while preserving the operation and concrete arguments. Feedback is used only to distinguish malformed outputs, inadmissible actions, and executable actions; the normalization rules are specified in Appendix~\ref{app:exact-context}.

Let $\mathcal I(c,v)$ contain all occurrences of normalized action $v$ in context $c$ within the current batch. We average their returns and compute the local advantage as
\begin{equation}
    Q_{\mathrm{loc}}(c,v)=
    \frac{1}{|\mathcal I(c,v)|}\sum_{j'\in\mathcal I(c,v)}G_{j'},
    \qquad
    A_j^{\mathrm{loc}}=
    \frac{Q_{\mathrm{loc}}(c_j,v_j)-\mu_{c_j}}{\sigma_{c_j}+\epsilon},
    \label{eq:concrete-advantage}
\end{equation}
where $\mu_c$ and $\sigma_c$ are the mean and standard deviation of all individual returns in context $c$, rather than of the action means. Thus, equivalent actions within a context receive the same advantage. Groups without distinct action means or return variation contribute zero local advantage.

\subsection{Task-Progress Advantage over Global State--Action Groups}
\label{sec:progress-advantage}

Matching current observations can mix histories with different task progress and bias the local baseline. Conditioning on history reduces this mismatch, but limited rollouts per task cannot reliably support detailed history groups. We therefore abstract histories and actions to pool comparable experience across tasks and training updates.

From the interaction prefix in Eq.~\ref{eq:interaction-history}, we maintain an execution state $B_{i,t}$ that records discovered entities, completed operations, and current relations:
\begin{equation}
    B_{i,t+1}=\operatorname{Update}(B_{i,t},a_{i,t},e_{i,t}).
    \label{eq:history-state}
\end{equation}
Updates incorporate new observations and accepted action effects. This state is used for credit estimation without changing the policy input.

Using predefined rules for each environment, we map entities to goal roles, preserving task family, predicate completion, and relations governing unfinished conditions whose prerequisites are satisfied. Since concrete commands would still yield sparsely observed alternatives, we also abstract their function:
\begin{equation}
    z_j=\phi_{\mathrm{prog}}(g_i,B_j),
    \qquad u_j=\psi_{\mathrm{prog}}(a_j;g_i,B_j).
    \label{eq:progress-mapping}
\end{equation}
Unlike $v_j$, $u_j$ can group commands involving different entities, such as acquiring different target objects. Identical $(z,u)$ pairs share statistics, while different $u$ under the same $z$ form the comparison group. Both maps are applied automatically; credit values are estimated from rollout returns. Appendix~\ref{app:progress-abstraction} specifies these mappings.

To account for differences in context difficulty and prevent repeated visits from dominating, we compute the mean residual within each $(c,z,u)$:
\begin{equation}
    m_{c,z,u}=\operatorname{Mean}
    \{\delta_j:c_j=c,\;z_j=z,\;u_j=u\},
    \qquad \delta_j=G_j-\mu_{c_j}.
    \label{eq:context-balanced-progress}
\end{equation}
Here $\delta_j$ measures the return relative to its concrete context's baseline. Each context contributes one such mean per $(z,u)$ to the global memory in each batch. Let $\mathcal E_k(z,u)$ be the multiset of these observations collected at update $k$ from contexts with multiple shareable action functions. We accumulate them in a global memory while discounting statistics from older policies:
\begin{equation}
    \begin{aligned}
        W^{(k)}_{z,u}
        &=\lambda_{\mathrm{ema}}W^{(k-1)}_{z,u}+|\mathcal E_k(z,u)|,\\
        S^{(k)}_{z,u}
        &=\lambda_{\mathrm{ema}}S^{(k-1)}_{z,u}
          +\sum_{m\in\mathcal E_k(z,u)}m,\\
        M^{(k)}_{z,u}
        &=\lambda_{\mathrm{ema}}M^{(k-1)}_{z,u}
          +\sum_{m\in\mathcal E_k(z,u)}m^2.
    \end{aligned}
    \label{eq:progress-memory}
\end{equation}
Here $0<\lambda_{\mathrm{ema}}<1$, and $W,S,M$ denote the decayed weight, weighted sum of residuals, and weighted sum of squared residuals. After incorporating the current batch, we query $q^{(k)}(z,u)=S^{(k)}_{z,u}/W^{(k)}_{z,u}$. Distinct context contributions are tracked separately to estimate effective support. Support depends on the balance of accumulated weight across distinct contexts, not on the number of recorded decisions. Thus, repeated observations from a few contexts do not by themselves justify a strong global advantage.

Let $\mathcal U_z$ contain the recorded action functions for state $z$. Omitting the update index, we compare each function with the group mean:
\begin{equation}
    \widetilde q(z,u)=q(z,u)
    -\frac{1}{|\mathcal U_z|}\sum_{u'\in\mathcal U_z}q(z,u').
    \label{eq:progress-contrast}
\end{equation}
Omitted task details and subsequent decisions still introduce variation, while support can differ substantially between action alternatives. We therefore scale each contrast by its support and uncertainty. With $n_{\mathrm{cmp}}(z,u)$ denoting the harmonic effective support of the action and its alternatives, we define $\alpha_{\mathrm{sup}}(z,u)=n_{\mathrm{cmp}}(z,u)/(n_{\mathrm{cmp}}(z,u)+\lambda_{\mathrm{sup}})$ and obtain
\begin{equation}
    A_j^{\mathrm{prog}}=
    \alpha_{\mathrm{sup}}(z_j,u_j)
    \frac{\widetilde q(z_j,u_j)}
    {\sqrt{\widehat\sigma_{z_j}^{2}+
    \widehat{\operatorname{SE}}_{\mathrm{diff}}^{2}(z_j,u_j)+\eta^2}}.
    \label{eq:progress-advantage}
\end{equation}
Here $\widehat\sigma_z^2$ measures dispersion across action means, $\widehat{\operatorname{SE}}_{\mathrm{diff}}$ estimates contrast uncertainty, and $\eta>0$ provides a numerical floor. The signal is zero unless the action is recorded and its group has at least two action functions, nonzero dispersion, and effective context support of at least two. Appendix~\ref{app:reliability} details the support and variance estimators. For WebShop, the resulting advantages are further centered over active decisions sharing $z$ within the current batch.

\subsection{Beyond-Local Policy Optimization}
\label{sec:blpo-objective}

We combine trajectory credit with the available local and task-progress advantages:
\begin{equation}
    A_j=A_i^{\mathrm{traj}}+\lambda_{\mathrm{step}}A_j^{\mathrm{step}},
    \qquad
    A_j^{\mathrm{step}}=\alpha_j A_j^{\mathrm{loc}}
    +(1-\alpha_j)A_j^{\mathrm{prog}}.
    \label{eq:final-advantage}
\end{equation}
Here $\alpha_j$ balances local and task-progress advantages according to the availability of the two estimators, while $\lambda_{\mathrm{step}}$ controls the contribution of step credit relative to trajectory credit. The weighting rule and parameter settings are specified in Appendix~\ref{app:hyperparameters}.

We assign $A_j$ to every token of action $a_j^{\mathrm{raw}}$ and maximize the clipped objective:
\begin{equation}
    \mathcal J_{\mathrm{BLPO}}(\theta)=
    \mathbb E_{j,\ell}\!\left[
    \min\!\left(\rho_{j,\ell}A_j,
    \operatorname{clip}(\rho_{j,\ell},1-\epsilon_{\mathrm{clip}},1+\epsilon_{\mathrm{clip}})A_j\right)
    -\beta\mathcal D_{\mathrm{KL},j,\ell}\right].
    \label{eq:objective}
\end{equation}
For action token $y_{j,\ell}$, $\rho_{j,\ell}=\frac{\pi_\theta(y_{j,\ell}\mid x_j,y_{j,<\ell})}{\pi_{\theta_{\mathrm{old}}}(y_{j,\ell}\mid x_j,y_{j,<\ell})}$ is the policy probability ratio. The parameter $\epsilon_{\mathrm{clip}}$ controls clipping, while $\mathcal D_{\mathrm{KL},j,\ell}$ regularizes the policy against a fixed reference with weight $\beta$. The update reuses sampled trajectories without a learned critic, process reward model, or additional environment interactions.

\section{Experiments}

\subsection{Experimental Setup}
\label{sec:experimental-setup}

\begin{table}[!b]
\centering
\caption{Performance on ALFWorld and WebShop. In-Success and Out-Success measure in-distribution and out-of-distribution success. WebShop Task Score and Task Success measure partial and complete request satisfaction. RL results are mean $\pm$ standard deviation over three seeds. Bold marks the best result per backbone; dashes indicate unavailable results.}
\label{tab:main-results}
\small
\setlength{\tabcolsep}{4.75pt}
\renewcommand{\arraystretch}{1.20}
\newcommand{\maincell}[2]{#1\textsubscript{\(\pm\)#2}}
\newcommand{\bestcell}[2]{\textbf{#1}\textsubscript{\(\pm\)#2}}
\begin{tabular}{cll|cccc}
\toprule
\multirow{2}{*}{Model} & \multicolumn{1}{c}{\multirow{2}{*}{Type}} & \multicolumn{1}{c|}{\multirow{2}{*}{Method}}
& \multicolumn{2}{c}{\textbf{ALFWorld}} & \multicolumn{2}{c}{\textbf{WebShop}} \\
& & & In-Success & Out-Success & Task Score & Task Success \\
\midrule
\multirow{2}{*}{Closed} & Prompting & GPT-4o & 48.0 & 46.0 & 31.8 & 23.7 \\
& Prompting & Gemini-2.5-Pro & 60.3 & 50.5 & 42.5 & 35.9 \\
\midrule
\multirow{9}{*}[-2pt]{\rotatebox{90}{\textit{Qwen2.5-1.5B-Instruct}}}
& Prompting & Qwen2.5 & 4.1 & -- & 23.1 & 5.2 \\
& Prompting & ReAct & 12.8 & -- & 40.1 & 11.3 \\
& Prompting & Reflexion & 21.8 & -- & 55.8 & 21.9 \\
& RL Training & PPO & \maincell{54.4}{3.1} & -- & \maincell{73.8}{3.0} & \maincell{51.5}{2.9} \\
& RL Training & RLOO & \maincell{69.7}{2.5} & \maincell{68.7}{10.7} & \maincell{73.9}{5.6} & \maincell{52.1}{6.7} \\
& RL Training & GRPO & \maincell{72.8}{3.6} & \maincell{70.1}{2.5} & \maincell{75.8}{3.5} & \maincell{56.8}{3.8} \\
\cmidrule(l){2-7}
& RL Training & GiGPO & \maincell{90.16}{0.22} & \maincell{84.76}{2.83} & \maincell{84.95}{2.57} & \maincell{66.53}{1.92} \\
& RL Training & HGPO & \maincell{92.77}{1.08} & \maincell{90.16}{0.78} & \maincell{85.56}{2.86} & \maincell{71.54}{4.00} \\
\rowcolor{black!5}
\cellcolor{white} & RL Training & \textbf{BLPO} & \bestcell{94.01}{0.23} & \bestcell{91.54}{0.60} & \bestcell{87.43}{0.56} & \bestcell{74.02}{1.37} \\
\midrule
\multirow{9}{*}[-2pt]{\rotatebox{90}{\textit{Qwen2.5-7B-Instruct}}}
& Prompting & Qwen2.5 & 14.8 & -- & 26.4 & 7.8 \\
& Prompting & ReAct & 31.2 & -- & 46.2 & 19.5 \\
& Prompting & Reflexion & 42.7 & -- & 58.1 & 28.8 \\
& RL Training & PPO & \maincell{77.08}{1.12} & \maincell{76.23}{1.46} & \maincell{81.4}{3.1} & \maincell{68.7}{5.1} \\
& RL Training & RLOO & \maincell{77.86}{0.03} & \maincell{73.95}{0.05} & \maincell{80.3}{3.2} & \maincell{65.7}{4.0} \\
& RL Training & GRPO & \maincell{78.64}{0.73} & \maincell{76.82}{1.47} & \maincell{79.3}{2.8} & \maincell{66.1}{3.7} \\
\cmidrule(l){2-7}
& RL Training & GiGPO & \maincell{93.29}{0.40} & \maincell{92.18}{0.39} & \maincell{88.93}{1.49} & \maincell{77.60}{1.68} \\
& RL Training & HGPO & \maincell{95.44}{0.62} & \maincell{92.05}{0.22} & \bestcell{88.96}{1.04} & \maincell{78.51}{1.40} \\
\rowcolor{black!5}
\cellcolor{white} & RL Training & \textbf{BLPO} & \bestcell{97.14}{1.33} & \bestcell{92.97}{0.68} & \maincell{88.90}{1.26} & \bestcell{79.43}{1.57} \\
\bottomrule
\end{tabular}
\end{table}

\paragraph{Benchmarks.}
Following prior work, we evaluate on ALFWorld~\citep{shridhar2021alfworldaligningtextembodied} and WebShop~\citep{yao2023webshopscalablerealworldweb} for long-horizon household and online shopping tasks. ALFWorld measures compositional manipulation success; WebShop evaluates purchases against user-specified product constraints.

\paragraph{Baselines.}
Baselines include GPT-4o, Gemini-2.5-Pro, direct Qwen2.5 prompting, ReAct~\citep{yao2023reactsynergizingreasoningacting}, and Reflexion~\citep{shinn2023reflexionlanguageagentsverbal}. We also compare PPO~\citep{schulman2017proximalpolicyoptimizationalgorithms}, RLOO~\citep{ahmadian2024backbasicsrevisitingreinforce}, GRPO~\citep{shao2024deepseekmathpushinglimitsmathematical}, GiGPO~\citep{feng2025groupingrouppolicyoptimizationllm}, and HGPO~\citep{he2026hierarchyofgroupspolicyoptimizationlonghorizon}. Table~\ref{tab:main-results} uses the baseline results reported by \citet{he2026hierarchyofgroupspolicyoptimizationlonghorizon}, with GiGPO and HGPO at history length $K=2$ to match our policy inputs.

\paragraph{Implementation details.}
Training uses Qwen2.5-1.5B-Instruct and Qwen2.5-7B-Instruct~\citep{qwen2025qwen25technicalreport}, 160 updates, 16 tasks per update, and eight trajectories per task. We measure learning curves and computational costs in our own runs. Appendix~\ref{app:implementation} provides evaluation and prompt details, with training and interaction parameters consolidated in Appendix~\ref{app:hyperparameters}.

\Needspace{4\baselineskip}
\raggedbottom
\subsection{Main Results}

\paragraph{\method{} achieves the best overall performance.}
Table~\ref{tab:main-results} reports the highest mean success for \method{} on both benchmarks at both scales. With Qwen2.5-1.5B, ALFWorld Out-Success rises from HGPO's 90.16\% to 91.54\%, and WebShop Task Success from 71.54\% to 74.02\%. Under matched rollout and update budgets, the 1.5B policy also surpasses 7B GRPO on all four metrics.

\paragraph{Global experience complements local credit assignment.}
GiGPO distinguishes decisions within trajectories; HGPO further conditions comparisons on history. \method{} pools decisions with shared task progress and action function across tasks and updates. Its gains over HGPO suggest that broader support, with context adjustment and reliability weighting, complements history conditioning within individual tasks.

\paragraph{Scaling and generalization.}
With the 7B backbone, \method{} reaches 97.14\% In-Success and 92.97\% Out-Success on ALFWorld, exceeding both GiGPO and HGPO. On WebShop, it attains a Task Score of 88.90, comparable to HGPO's 88.96, while increasing Task Success from 78.51\% to 79.43\%. Thus, \method{} improves complete task fulfillment even when partial request satisfaction remains similar.

\subsection{Ablation Study}

\begin{wraptable}{r}{7.3cm}
\centering
\vspace{-4mm}
\fontsize{8}{9.3}\selectfont
\caption{Component ablations with Qwen2.5-1.5B. Gray values show absolute drops from full \method{}.}
\vspace{-2mm}
\label{tab:ablation}
\fontsize{8.3}{9.6}\selectfont
\setlength{\tabcolsep}{0pt}
\renewcommand{\arraystretch}{1.00}
\newlength{\blpoMetricWidth}
\setlength{\blpoMetricWidth}{\dimexpr(\linewidth-52pt)/4\relax}
\newcommand{\ablcell}[2]{#1\textsubscript{\color{black!50}\fontsize{5.5}{6}\selectfont$\blacktriangledown$\,#2}}
\begin{tabular}{@{}p{52pt}*{4}{>{\centering\arraybackslash}p{\blpoMetricWidth}}@{}}
\toprule
& \multicolumn{2}{c}{ALFWorld} & \multicolumn{2}{c}{WebShop} \\
\cmidrule(lr){2-3}\cmidrule(l){4-5}
Variant & In-Suc. & Out-Suc. & Score & Success \\
\midrule
\rowcolor{black!6}
\textbf{\method{}} & \textbf{94.01} & \textbf{91.54} & \textbf{87.43} & \textbf{74.02} \\
\addlinespace[1.5pt]
\multicolumn{5}{@{}l}{\color{black!65}\fontsize{7.5}{9}\selectfont\textit{Local credit}} \\
w/o Act. pool.
& \ablcell{91.34}{2.67} & \ablcell{83.14}{8.40} & \ablcell{82.79}{4.64} & \ablcell{66.21}{7.81} \\
\addlinespace[1.5pt]
\multicolumn{5}{@{}l}{\color{black!65}\fontsize{7.5}{9}\selectfont\textit{Task-progress groups}} \\
w/o $A^{\mathrm{prog}}$
& \ablcell{92.25}{1.76} & \ablcell{83.53}{8.01} & \ablcell{82.75}{4.68} & \ablcell{68.49}{5.53} \\
w/o Act. abstr.
& \ablcell{88.80}{5.21} & \ablcell{82.94}{8.60} & \ablcell{86.83}{0.60} & \ablcell{71.22}{2.80} \\
\addlinespace[1.5pt]
\multicolumn{5}{@{}l}{\color{black!65}\fontsize{7.5}{9}\selectfont\textit{Global estimation}} \\
w/o Residual
& \ablcell{88.41}{5.60} & \ablcell{81.97}{9.57} & \ablcell{85.14}{2.29} & \ablcell{73.24}{0.78} \\
w/o EMA
& \ablcell{90.89}{3.12} & \ablcell{84.11}{7.43} & \ablcell{86.09}{1.34} & \ablcell{69.92}{4.10} \\
w/o Conf.
& \ablcell{89.97}{4.04} & \ablcell{81.05}{10.49} & \ablcell{85.00}{2.43} & \ablcell{71.09}{2.93} \\
\bottomrule
\end{tabular}
\vspace{-5mm}
\end{wraptable}

\paragraph{Local credit.}
Table~\ref{tab:ablation} shows that removing action-level pooling reduces ALFWorld Out-Success by 8.40 points and WebShop Task Success by 7.81 points. Using individual returns allows different continuations to assign inconsistent local credit to the same action.

\paragraph{Task-progress groups.}
The loss of task-progress credit lowers WebShop Task Score from 87.43 to 82.75 and Task Success from 74.02\% to 68.49\%. When local comparisons are uninformative, trajectory credit alone cannot distinguish useful steps from later failures. Removing action abstraction also lowers success in both environments despite retaining progress states and global memory. Concrete labels split equivalent decisions across objects and tasks, limiting comparison support.

\paragraph{Global estimation.}
Removing residualization or confidence adjustment lowers ALFWorld Out-Success below the variant without task-progress credit, supporting control of context bias and noise in shared estimates. On WebShop, removing confidence adjustment lowers Task Success by 2.93 points; removing EMA lowers it by 4.10 points. The latter uses only the current batch, supporting temporal accumulation for sparse action groups.

\Needspace{11\baselineskip}
\subsection{Further Analysis}

\noindent\begin{minipage}{\textwidth}
\setlength{\intextsep}{0pt}
\begin{wrapfigure}[10]{r}{7.4cm}
\centering
\includegraphics[width=\linewidth]{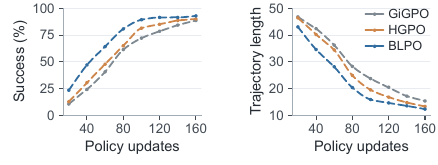}
\caption{ALFWorld learning dynamics.}
\label{fig:training-dynamics}
\end{wrapfigure}
\noindent\textbf{Learning dynamics.}\quad
\method{} learns faster and uses fewer actions in Figure~\ref{fig:training-dynamics}. At update 80, success reaches 80.7\% with 20.4 actions per task, versus HGPO's 65.0\% and 24.9 actions. \method{} retains the highest final success and shortest trajectories. Success rises as the interaction count falls, indicating that shorter trajectories do not come at the expense of task completion. Later gains in success are smaller for all methods, but trajectories continue to shorten. Length averages cover all 140 tasks, including failures and timeouts, so the reduction reflects lower overall interaction cost across the complete set of evaluation tasks.
\par\end{minipage}

\Needspace{12\baselineskip}
\par\noindent\begin{minipage}{\textwidth}
\setlength{\intextsep}{0pt}
\begin{wraptable}[10]{r}{7.4cm}
\centering
\caption{Training cost on ALFWorld (4 A100s, 160 updates). Adv.\ and Update are mean times per update; Share is their ratio. Evaluation time is excluded.}
\label{tab:efficiency}
\fontsize{8}{9.3}\selectfont
\setlength{\tabcolsep}{1pt}
\renewcommand{\arraystretch}{1.12}
\begin{tabular*}{\linewidth}{@{\extracolsep{\fill}}ccccc@{}}
\toprule
Method & Adv.\ (s) & Update (s) & Share & Total (h) \\
\midrule
GiGPO & 0.89 & 335.2 & 0.26\% & 14.90 \\
HGPO  & 0.96 & 362.6 & 0.26\% & 16.11 \\
\textbf{\method{}} & 2.12 & 349.7 & 0.61\% & 15.54 \\
\bottomrule
\end{tabular*}
\end{wraptable}
\noindent\textbf{Computational cost.}\quad
Global credit adds little cost in Table~\ref{tab:efficiency}. Computing advantages takes 2.12 seconds per update, or 0.61\% of update time. The increase over GiGPO and HGPO is approximately 1.2 seconds per update. This cost includes local grouping, memory maintenance, and reliability estimation. Memory retains aggregated statistics rather than replaying past trajectories, so queries reuse historical evidence without reevaluating earlier rollouts. Maintaining this broader comparison pool requires neither a separate value network nor additional environment interactions. The full run takes 15.54 hours, within the baseline range. Overall time also depends on the trajectories produced by the policy, not just the estimator's arithmetic. Appendix~\ref{app:efficiency} gives timing details.
\par\end{minipage}

\begin{figure}[!htbp]
    \centering
    \includegraphics[width=\textwidth]{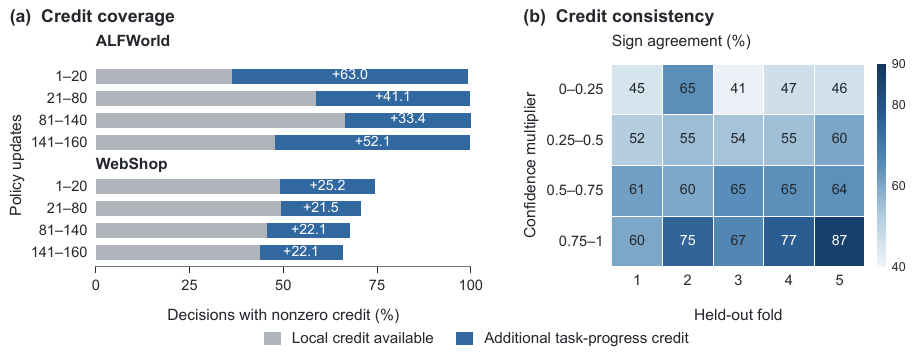}
    \caption{Coverage and consistency of global credit. (a) Fractions of decisions with nonzero local credit (gray) and additional coverage from task-progress credit (blue). (b) Sign agreement across confidence bins and held-out task folds in ALFWorld.}
    \label{fig:global-credit}
    \label{fig:credit-reliability}
\end{figure}

\paragraph{Coverage of task-progress credit.}
Figure~\ref{fig:global-credit}a shows that task-progress credit remains available where local advantages are zero. During updates 141--160, it covers 52.1\% of ALFWorld decisions and 22.1\% of WebShop decisions. Local advantages can vanish when available rollouts do not distinguish alternatives within a concrete context. These decisions retain trajectory credit, but the common outcome does not distinguish their intermediate contributions. Task-progress groups supply comparisons across tasks, so preferences can remain available when the current task offers little local contrast. This additional coverage persists late in training rather than merely rescaling nonzero local signals. Appendix~\ref{app:credit-data} defines the diagnostics.

\Needspace{4\baselineskip}
\paragraph{Reliability of global credit.}
Figure~\ref{fig:credit-reliability}b compares preferences estimated from four task subsets with residual contrasts in the held-out subset. Held-out tasks do not contribute to the estimates, so agreement tests preferences across task instances rather than their fit to the observations used for estimation. This probes whether progress-aligned groups preserve action preferences despite differences in the concrete tasks. The highest confidence bin achieves greater sign agreement than the lowest in all five folds. Agreement is not monotonic in every fold, but the overall pattern supports reliability weighting of global preferences. Appendix~\ref{app:offline-credit} gives the protocol.

\raggedbottom
\begin{table}[!htbp]
\centering
\begin{minipage}[t]{0.48\textwidth}
\vspace{0pt}
\caption{Qwen2.5-1.5B results under global sharing, same-task-only sharing, and removal of the task-progress credit branch.}
\label{tab:cross-task-experience}
\small
\setlength{\tabcolsep}{3pt}
\renewcommand{\arraystretch}{1.50}
\begin{tabular*}{\linewidth}{@{\extracolsep{\fill}}lcccc@{}}
\toprule
& \multicolumn{2}{c}{ALFWorld} & \multicolumn{2}{c}{WebShop} \\
\cmidrule(lr){2-3}\cmidrule(l){4-5}
Setting & In-Suc. & Out-Suc. & Score & Success \\
\midrule
\textbf{Full \method{}} & \textbf{94.01} & \textbf{91.54} & \textbf{87.43} & \textbf{74.02} \\
Same-task-only & 87.63 & 81.97 & 87.19 & 66.99 \\
w/o $A^{\mathrm{prog}}$ & 92.25 & 83.53 & 82.75 & 68.49 \\
\bottomrule
\end{tabular*}
\end{minipage}\hfill
\begin{minipage}[t]{0.48\textwidth}
\vspace{0pt}
\centering
\includegraphics[width=\linewidth,trim=0 11 0 3,clip]{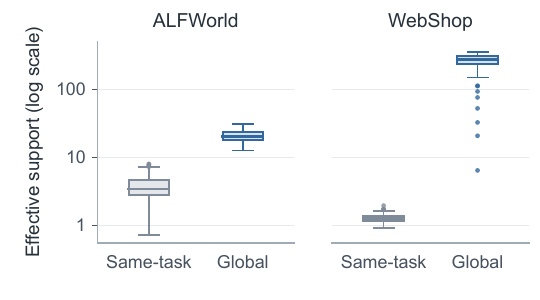}
\captionof{figure}{Effective context support from same-task and global experience. Boxes summarize per-update means in full \method{}.}
\label{fig:cross-task-support}
\end{minipage}
\end{table}

\Needspace{2\baselineskip}
\paragraph{Benefit of cross-task experience.}
Same-task-only restricts memory to each task instance, retaining the abstraction, cross-update accumulation, and all other estimation and fusion rules. Table~\ref{tab:cross-task-experience} shows lower success than both full \method{} and removing task-progress credit. On WebShop, its Task Score stays near full \method{}, but Task Success falls by 7.03 points, separating partial satisfaction from completion. With the abstraction fixed, these findings support cross-task experience beyond the shared domain prior. Removing the branch also performs better than restricting it to one task, so retaining task-progress credit alone does not explain the gains. Its benefit depends on the comparison evidence supplied to the estimator; limited support can yield noisy estimates that interfere with local credit.

\Needspace{5\baselineskip}
\paragraph{Effective context support.}
Figure~\ref{fig:cross-task-support} shows greater effective context support with cross-task sharing, especially in WebShop. Effective support accounts for the concentration of contribution weights across contexts, rather than simply counting recorded decisions. Repeated visits to one context add weight, not distinct contexts. Memory can therefore persist across updates yet provide narrow support, as the same-task WebShop estimates illustrate. The reliability correction uses support on both sides of each action comparison. A broader pool can strengthen weakly supported action comparisons, helping explain the value of global credit under the same abstraction.

\Needspace{17\baselineskip}
\begingroup
\setlength{\intextsep}{0pt}
\begin{wrapfigure}[16]{r}{0.45\textwidth}
    \centering
    \includegraphics[width=\linewidth]{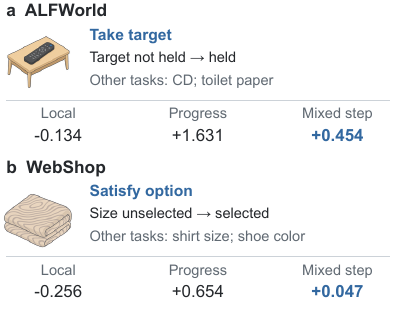}
    \caption{Examples of cross-task credit in ALFWorld and WebShop.}
    \label{fig:credit-cases}
\end{wrapfigure}
\hyphenpenalty=500
\noindent\textbf{Case study.}\quad
Figure~\ref{fig:credit-cases} shows two actions receiving negative local credit despite advancing necessary subgoals: acquiring the target object in \mbox{ALFWorld} and selecting the requested size in \mbox{WebShop}. In the first case, taking a remote control shares its function with acquiring other target objects, including CDs and toilet paper, in placement tasks. In the second, selecting a blanket size is compared with satisfying required options for other products, including shirt sizes and shoe colors. Although the WebShop trajectory ultimately fails, selecting the requested size still resolves a necessary constraint, separating this useful step from the final outcome. Evidence from these groups turns the mixed step advantage positive in both cases, aligning credit with the actions' contribution to task progress. The global preferences remain positive after excluding the current task's contributions, confirming support from other tasks.
\par\endgroup

\Needspace{4\baselineskip}
\section{Discussion and Limitations}

Our results support global credit estimation that controls context difficulty and uncertainty, rather than increasing comparison support alone. The current instantiation uses environment-specific rules to align task progress and action functions, while sharing the statistical estimator across environments. Applying \method{} to new domains calls for suitable comparison groups; learning transferable abstractions is a promising direction for broadening its applicability.

\section{Conclusion}

We introduced \method{} to estimate task-progress advantages from global experience while controlling context bias and uncertainty. Combined with trajectory and local credit, these advantages improve task success on ALFWorld and WebShop at both model scales. Faster learning and shorter ALFWorld trajectories further support global experience as a useful source of decision-level supervision.

\bibliography{iclr2027_conference,additional_references}
\bibliographystyle{iclr2027_conference}

\appendix
\setlength{\parskip}{4pt}
\setlength{\intextsep}{6pt plus 1pt minus 1pt}
\captionsetup{skip=6pt}
\raggedbottom

\section{Implementation Details}
\label{app:implementation}

\subsection{Evaluation protocol}

Our runs share the same environment interface, prompts, action parser, and interaction limits. Following GiGPO~\citep{feng2025groupingrouppolicyoptimizationllm}, we report success rates on ALFWorld and both task scores and success rates on WebShop. We evaluate each final checkpoint with three evaluation seeds. For each seed, ALFWorld uses 512 episodes on each of the In and Out splits, and WebShop uses 512 tasks. We compute each metric separately for each seed and report the mean and sample standard deviation across seeds. For the learning curves in Figure~\ref{fig:training-dynamics}, all methods and checkpoints are evaluated on the same 140 ALFWorld In tasks.

\subsection{Rollout reward convention}
\label{app:rollout-reward}

Our training runs use the same rule-based rollout reward. Following GiGPO~\citep{feng2025groupingrouppolicyoptimizationllm}, successful task completion receives a reward of 10 and failure receives 0; a format-invalid model output incurs an additional penalty:
\begin{equation}
    r_{i,t}=r^{\mathrm{env}}_{i,t}
    -\lambda_{\mathrm{fmt}}
    \mathbb{I}\!\left[a^{\mathrm{raw}}_{i,t}\text{ is malformed}\right],
    \qquad \lambda_{\mathrm{fmt}}=0.1.
    \label{eq:rollout-reward}
\end{equation}
The penalty is applied before both the episodic return in Equation~\ref{eq:trajectory-advantage} and the discounted future return in Equation~\ref{eq:step-return} are computed. It applies only when the raw output violates the prescribed response format or cannot be parsed into an action. A well formed but inadmissible action is recorded separately and is not treated as a format violation.

\subsection{Exact context and local action pooling}
\label{app:exact-context}

The concrete context in Equation~\ref{eq:concrete-mapping} combines the task rollout-group identifier $b_i$ with the action's pre-execution observation. Two decisions enter the same local group only when both components match. Observation matching preserves the complete content, including concrete entities and arguments, without entity replacement or semantic matching. Thus, identical observations from different task groups remain separate, as do observations that differ in their concrete objects. Earlier interaction history and WebShop's additional public-state fields are not used for local grouping.

For action normalization, we apply Unicode NFKC normalization to the extracted action text, trim its ends, convert letters to lowercase, and collapse consecutive whitespace. This removes differences in writing without deleting the operation, entity identifiers, or arguments. The normalized text is paired with one of three validity classes: malformed output, a well formed but inadmissible action, or an executable action. Validity is determined from the parser and environment feedback after execution; it does not change the pre-action observation used for grouping. For example, case and whitespace variants of \texttt{take apple 1 from table 1} represent the same normalized action. Changing the object identifier or validity class instead produces a distinct normalized action.

Once the rollout is complete, discounted returns are averaged within each $(c,v)$ group. The baseline $\mu_c$ and sample standard deviation $\sigma_c$ are instead computed over all individual decision occurrences in $c$, not over equally weighted action means. The pooled advantage is assigned to every occurrence of the same action. A group is inactive when its return variation is negligible or all pooled action means coincide with the baseline, including the single-action case. All local groups and statistics are rebuilt for each rollout batch. The w/o action pooling variant replaces $Q_{\mathrm{loc}}(c_j,v_j)$ with $G_j$ in Equation~\ref{eq:concrete-advantage}, retaining the context baseline and scale. Its activity depends on individual-return variation rather than differences between action means. For the global residual in Equation~\ref{eq:context-balanced-progress}, we subtract $\mu_c$ from each occurrence's own return $G_j$, not from its pooled action value.

\subsection{History and task-progress abstraction}
\label{app:progress-abstraction}

\paragraph{Construction principle.}
The rules follow a common construction principle across the two environments. We begin with task completion conditions and identify observable facts that track whether their prerequisites have been met. Task descriptions define goal-relative roles, while interaction history and environment feedback update the relevant facts before each decision. States summarize task progress, and actions are mapped to their functions relative to the goal and current progress. This retains distinctions such as acquiring an unplaced target versus retaking a placed one, while allowing object identities to differ. The environment-specific rules are reused across task instances and executed automatically, without annotating individual trajectories. They define comparison groups rather than numerical advantages or additional progress rewards; action preferences are estimated from rollout returns. State construction uses only pre-action information and does not extend the policy input.

\paragraph{ALFWorld history and progress states.}
Task metadata identifies the task family, target object, destination, and required tool. Placement tasks define a destination relation; Heat, Cool, and Clean add a required property; Pick2 requires two placements; Look requires inspection under light. We reconstruct the relevant facts by replaying accepted actions, tracking possession, object locations, completed transformations, placement counts, encountered entities, and container or tool states. Rejected actions leave these facts unchanged, and removing a previously placed instance decreases the placement count.

The task-progress state combines the task family, the types of goal conditions, their progress representations, and their joint completion pattern. Each condition retains goal-relative relations, a coarse progress stage, and relevant facts such as whether the target is held or the required tool is accessible. Transformation stages distinguish an undiscovered target, discovery or possession, possession with a seen or accessible tool, placement in the tool, and completed transformation. Placement conditions separately retain whether delivery remains unfinished; Pick2 additionally records placement count, while Look retains target and lighting conditions. These stages are categorical grouping features and do not provide additional rewards.

\paragraph{ALFWorld action functions.}
Commands are parsed into operations and arguments, with entities mapped to task roles such as target, destination, and tool. Functions distinguish navigation, acquisition, transformation, placement, inspection, and exploration. For Pick2, taking a target from the destination is classified as a retake rather than acquisition, separating actions that can undo completed placement.

\paragraph{WebShop decision frontiers.}
Fixed patterns extract options and price limits; normalized tokens represent the requested product. These constraints are compared with the observation and public page state, including query, selected options, clickable elements, and prior interaction counts. The task-progress state retains only the page type and active decision frontier, omitting concrete query and product identities.

Frontier rules are evaluated in order. On a results page, an affordable candidate covering at least two-thirds of the target product tokens marks readiness to inspect a candidate. Otherwise, query coverage below two-thirds indicates a need for query refinement. If coverage is sufficient and another results page exists, the frontier instead indicates further result exploration. On an item page, excessive price or zero product-token overlap first marks the item as unsuitable. Missing or conflicting options then indicate a need for option selection when a requested choice is available; otherwise, the item is unsuitable. Once required options are satisfied, sufficient token coverage and an acceptable price mark readiness to purchase. Remaining cases indicate a need for information only if product details have not yet been inspected; otherwise, the frontier remains unresolved. An absent price limit imposes no price restriction. These checks use current public information, not a hidden purchase-success label.

\paragraph{WebShop action functions.}
Action functions distinguish opening a strongly matching, affordable candidate from selecting a requested option that remains unmet. Conflicting or repeated selections are separated, as are inspection, purchase, backtracking, and result exploration. Table~\ref{tab:progress-grounding} illustrates these functional distinctions; sharing requires both the abstract state and action function to match.

\begin{table}[!htbp]
\centering
\caption{Examples of task-relative distinctions preserved by action abstraction.}
\label{tab:progress-grounding}
\small
\setlength{\tabcolsep}{5pt}
\begin{tabular}{@{}p{0.14\linewidth}p{0.46\linewidth}p{0.34\linewidth}@{}}
\toprule
Environment & Task-relative decision & Action function \\
\midrule
ALFWorld & {\raggedright In Pick2, take a target from outside the destination.\par} & {\raggedright Acquire an unplaced target\par} \\
ALFWorld & {\raggedright In Pick2, take a target back out of the destination.\par} & {\raggedright Retake a placed target\par} \\
WebShop & {\raggedright Select a requested option that is not yet satisfied.\par} & {\raggedright Satisfy an option\par} \\
WebShop & {\raggedright Select a conflicting option or repeat an already satisfied choice.\par} & {\raggedright Violate or repeat an option\par} \\
\bottomrule
\end{tabular}
\end{table}

\paragraph{Sharing boundary.}
WebShop excludes unresolved frontiers and action functions from global sharing, including open-ended search queries and results whose product titles and prices cannot be aligned reliably. Item subpages are also excluded because they generally offer only a return action. Parser and execution-validity checks exclude malformed or inadmissible commands without altering the pre-action state; excluded decisions retain access to trajectory or local credit. These gates determine whether sharing is allowed; Appendix~\ref{app:reliability} describes how support and uncertainty scale the resulting estimate. The grouping interface and estimator are shared across environments, while task parsing, tracked facts, and functional mappings remain environment-specific.

\subsection{Global memory and reliability}
\label{app:global-memory}
\label{app:reliability}

This section specifies the memory maintenance and reliability estimators used in Section~\ref{sec:progress-advantage}. For each observed $(z,u)$ pair, the memory maintains the moments in Equation~\ref{eq:progress-memory}, the sum of squared observation weights, and the last update. It also tracks decayed contributions from contexts recognized across updates, separately from the batch-local grouping $c$. For this support calculation, ALFWorld matches the task family and observation, whereas WebShop matches the task instance and observation. This tracks repeated support across updates without changing local comparison groups.

Before a record is read or updated after $\Delta$ optimization steps, its moments and context weights are multiplied by $\lambda_{\mathrm{ema}}^{\Delta}$; the sum of squared weights is multiplied by $\lambda_{\mathrm{ema}}^{2\Delta}$. We set $\lambda_{\mathrm{ema}}=2^{-1/H}$ for a half-life of $H$ updates. A local context supplies new residual observations only when it contains at least two shareable occurrences with different abstract action functions. Repeated occurrences within each $(c,z,u)$ are then averaged as in Equation~\ref{eq:context-balanced-progress}, and each resulting mean adds one unit of weight. All eligible current-batch observations are written before the advantages for that update are queried.

Let $C_{z,u}(\chi)$ be the decayed contribution weight of context $\chi$ under this matching rule. The effective context support is
\begin{equation}
    n_{\mathrm{eff}}(z,u)=
    \frac{\bigl(\sum_\chi C_{z,u}(\chi)\bigr)^2}
         {\sum_\chi C_{z,u}(\chi)^2}.
    \label{eq:effective-context-support}
\end{equation}
We first take the harmonic mean of the effective supports of all alternatives to $u$, then take the harmonic mean of that value and $n_{\mathrm{eff}}(z,u)$ to obtain $n_{\mathrm{cmp}}(z,u)$. This makes the support correction in Equation~\ref{eq:progress-advantage} sensitive to the less-supported side of the comparison.

To compute uncertainty, let $D_{z,u}$ denote the stored sum of squared observation weights. The weighted residual variance and the estimated variance of the action mean are
\begin{equation}
    \widehat v(z,u)=
    \frac{\max\{M_{z,u}-S_{z,u}^2/W_{z,u},0\}}
         {W_{z,u}-D_{z,u}/W_{z,u}},
    \qquad
    h(z,u)=\frac{\widehat v(z,u)}{\max\{n_{\mathrm{eff}}(z,u),1\}}.
    \label{eq:memory-variance}
\end{equation}
We use zero variance when its denominator is numerically negligible. For $K=|\mathcal U_z|$ recorded action alternatives, the squared standard error of the centered contrast is
\begin{equation}
    \widehat{\operatorname{SE}}_{\mathrm{diff}}^2(z,u)
    =\left(\frac{K-1}{K}\right)^2 h(z,u)
    +\frac{1}{K^2}\sum_{u'\ne u}h(z,u').
    \label{eq:contrast-standard-error}
\end{equation}
The dispersion $\widehat\sigma_z^2$ is the sample variance of the recorded action means, and $\eta^2$ supplies the additive floor in Equation~\ref{eq:progress-advantage}.

A global group is active only when it has at least two action alternatives with non-negligible dispersion and effective context support of at least two. For this group-level support, a context's weight is the maximum of its weights across action alternatives, avoiding repeated counting within the group. A queried action must also have a memory record. After reliability correction, WebShop subtracts the current-batch mean over active decisions in each progress-state group; no further variance normalization is applied. The reported diagnostics and case values include this centering.

\Needspace{4\baselineskip}
\subsection{Training and interaction settings}
\label{app:hyperparameters}

We train for 160 policy updates, sampling 16 tasks and eight trajectories per task at each update, for a total of 128 trajectories. The policy learning rate is $10^{-6}$, with clipping threshold $\epsilon_{\mathrm{clip}}=0.2$ and KL coefficient $\beta=0.01$. We use a discount factor of $\gamma=0.95$ and format penalty $\lambda_{\mathrm{fmt}}=0.1$.

The global memory uses an EMA half-life of eight updates, with support shrinkage $\lambda_{\mathrm{sup}}=4$ and numerical floor $\eta=0.05$. For advantage fusion, $\alpha_j=2/3$ when both step estimators are active, $1$ for local credit only, and $0$ for task-progress credit only. If neither estimator is active, the update uses trajectory credit alone; we set $\lambda_{\mathrm{step}}=1$ throughout.

Interactions are limited to 50 steps in ALFWorld and 30 in WebShop. Training and evaluation sample responses at temperatures 1.0 and 0.4, respectively. The policy retains up to two recent observation--action pairs and generates at most 512 tokens per response. After chat formatting, prompts are left-truncated to 2,048 tokens for ALFWorld and 4,096 for WebShop.

\subsection{Computational profile}
\label{app:efficiency}

The computational measurements in Table~\ref{tab:efficiency} use four NVIDIA A100 40GB GPUs and cover all 160 ALFWorld policy updates. We average the logged advantage-computation and full-update times across updates, excluding validation and checkpoint evaluation. Advantage computation includes local grouping, memory maintenance, and reliability estimation. Full-update time also includes rollout generation and policy optimization, so its variation reflects both computational overhead and the trajectory lengths produced by each policy.

\Needspace{4\baselineskip}
\subsection{Prompt templates}
\label{app:prompts}

Both environments use the same prompt templates during training and evaluation. The environment prompt is passed as a single user message through the Qwen chat template, which adds the default system message and the assistant generation prefix. Each subsequent input includes the task, current observation, admissible actions, and recent history, subject to the limits in Appendix~\ref{app:hyperparameters}. History contains actions submitted to the environment, not the model's reasoning text. This policy input is separate from the compact history state maintained by the credit estimator.

\begingroup
\definecolor{PromptInk}{HTML}{2E465B}
\definecolor{PromptHeader}{HTML}{EAF0F5}
\definecolor{PromptRule}{HTML}{BCCAD6}
\newcommand{\promptslot}[1]{\textcolor{PromptInk}{\texttt{\{#1\}}}}
\newcommand{\prompttag}[1]{\texttt{#1}}
\long\def\promptpanel#1#2{%
  \begin{mdframed}[linecolor=PromptRule,linewidth=0.45pt,
    innerleftmargin=8pt,innerrightmargin=8pt,innertopmargin=6pt,innerbottommargin=6pt,
    skipabove=6pt,skipbelow=6pt,frametitle={#1},repeatframetitle=true,
    frametitlebackgroundcolor=PromptHeader,frametitlefont=\small\bfseries\color{PromptInk},
    splitbottomskip=6pt,splittopskip=6pt]
  \small\rmfamily\raggedright #2
  \end{mdframed}}

\Needspace{4\baselineskip}
\paragraph{ALFWorld.}
The template below is used after the initial decision. At the initial decision, its opening sentence is ``You are an expert agent operating in the ALFRED Embodied Environment.'' The task is already included in the initial observation, and the history and step counters are omitted. The admissible action list excludes \texttt{help}.

\promptpanel{ALFWorld prompt template}{%
You are an expert agent operating in the ALFRED Embodied Environment. Your task is to: \promptslot{task\_description}\par\medskip
Prior to this step, you have already taken \promptslot{step\_count} step(s). Below are the most recent \promptslot{history\_length} observations and the corresponding actions you took: \promptslot{action\_history}\par\medskip
You are now at step \promptslot{current\_step} and your current observation is: \promptslot{current\_observation}\par\smallskip
Your admissible actions of the current situation are: [\promptslot{admissible\_actions}].\par\medskip
Now it's your turn to take an action.\par\smallskip
You should first reason step-by-step about the current situation. This reasoning process MUST be enclosed within \prompttag{<think> </think>} tags.\par\smallskip
Once you've finished your reasoning, you should choose an admissible action for current step and present it within \prompttag{<action> </action>} tags.}

\Needspace{4\baselineskip}
\paragraph{WebShop.}
The task is extracted from the instruction field, and available actions are constructed from the search interface and clickable page elements. The initial decision uses the same instructions below but omits the history sentence and replaces ``You are now at step \{current\_step\} and your current observation is:'' with ``Your current observation is:''. History entries in both environments take the form \texttt{[Observation k: '...', Action k: '...']}.

\promptpanel{WebShop prompt template}{%
You are an expert autonomous agent operating in the WebShop e-commerce environment.\par\smallskip
Your task is to: \promptslot{task\_description}.\par\medskip
Prior to this step, you have already taken \promptslot{step\_count} step(s). Below are the most recent \promptslot{history\_length} observations and the corresponding actions you took: \promptslot{action\_history}\par\medskip
You are now at step \promptslot{current\_step} and your current observation is: \promptslot{current\_observation}.\par\smallskip
Your admissible actions of the current situation are: [\promptslot{available\_actions}].\par\medskip
Now it's your turn to take one action for the current step.\par\smallskip
You should first reason step-by-step about the current situation, then think carefully which admissible action best advances the shopping goal. This reasoning process MUST be enclosed within \prompttag{<think> </think>} tags.\par\smallskip
Once you've finished your reasoning, you should choose an admissible action for current step and present it within \prompttag{<action> </action>} tags.}
\endgroup

\section{Additional Experimental Diagnostics}
\label{app:additional-experiments}

\subsection{Data and statistical procedures}
\label{app:credit-data}

The credit analyses use 529,352 ALFWorld decisions and 162,121 WebShop decisions collected over 160 policy updates in each environment. Each decision is associated with its task, concrete context, task-progress state, action function, discounted return, and both credit estimates. Coverage is the proportion of decisions with a nonzero advantage from the corresponding estimator.

\subsection{Cross-task consistency test}
\label{app:offline-credit}

The test in Figure~\ref{fig:credit-reliability}b uses 3,473 supported action contrasts. Tasks are partitioned into five disjoint folds, keeping all records from each task together. Four folds estimate context-balanced residual action means, and the held-out fold provides comparison estimates for the same progress groups. Contributions retain their EMA weights at update 160. Both estimates are centered over common action alternatives, requiring at least two alternatives and effective context support of at least two per action on each side. We compare the signs of nonzero contrasts and bin their agreement by the confidence multiplier. Away from the numerical floor, this multiplier is
\begin{equation}
    \kappa(z,u)=\alpha_{\mathrm{sup}}(z,u)
    \frac{\widehat\sigma_z}
    {\sqrt{\widehat\sigma_z^2+
    \widehat{\operatorname{SE}}_{\mathrm{diff}}^2(z,u)+\eta^2}}.
    \label{eq:confidence-multiplier}
\end{equation}
It measures shrinkage relative to the unscaled standardized contrast $\widetilde q(z,u)/\widehat\sigma_z$, not a probability of correctness. Numerically, we divide the magnitude of the reliability-scaled contrast by that of the unscaled contrast, flooring the latter at $10^{-6}$ and capping the ratio at one. All quantities are recomputed from the estimation folds; agreement is summarized in four bins of width 0.25. Repeating the procedure for each held-out fold produces the five columns in the figure.

\Needspace{4\baselineskip}
\subsection{Complete decision-credit decomposition}
\label{app:credit-decomposition}

\begin{figure}[!htbp]
    \centering
    \includegraphics[width=\textwidth]{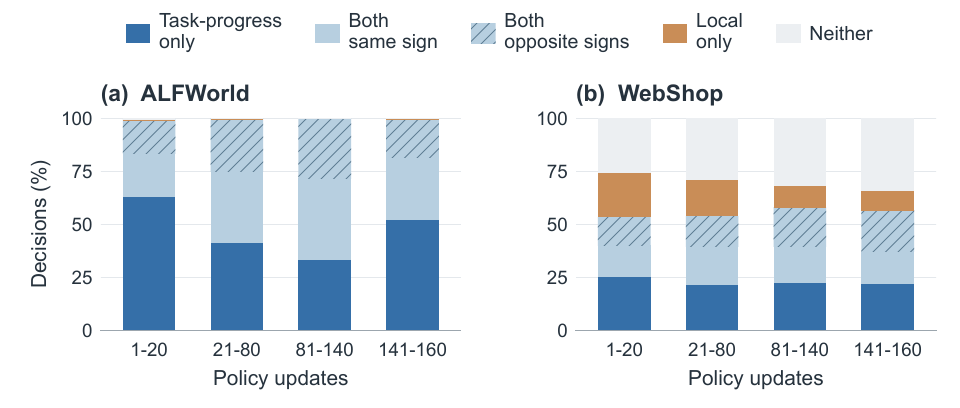}
    \caption{Local and task-progress credit across training phases. Bars partition all decisions by nonzero credit availability; hatching marks shared coverage with opposing advantage signs.}
    \label{fig:credit-coverage}
\end{figure}

Figure~\ref{fig:credit-coverage} shows that broader coverage is a persistent contribution of task-progress credit, rather than an effect confined to early exploration. Its exclusive coverage remains substantial in every phase, while local-only credit accounts for less than 0.1\% of ALFWorld decisions. In WebShop, task-progress-only coverage stays near 22\% after the initial phase, even as local-only coverage decreases. Where both estimators are active, their signs can differ because they compare actions against different reference groups. The global branch therefore changes both the availability and the direction of decision-level supervision; the sign decomposition measures these differences, not which estimator is correct.

\Needspace{4\baselineskip}
\subsection{Performance across ALFWorld task types}
\label{app:task-types}

\begingroup
\setlength{\intextsep}{6pt}
\begin{figure}[!htbp]
    \centering
    \begin{minipage}[c]{0.43\textwidth}
        \includegraphics[width=\linewidth]{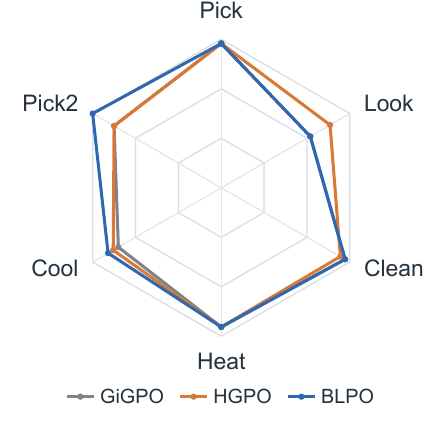}
    \end{minipage}\hfill
    \begin{minipage}[c]{0.53\textwidth}
        \raggedright
        Figure~\ref{fig:alfworld-task-types} shows \method{}'s largest gain on Pick2, which requires retaining completed placement progress while handling a second object. Task-progress groups can credit useful intermediate steps through comparisons with similar stages in other tasks, even when subsequent execution fails. Look instead requires target possession and lighting conditions to hold jointly. This greater dependence on the immediate configuration may offer less scope for progress-based sharing, providing a possible explanation for \method{}'s lower success than HGPO on Look.
    \end{minipage}
    \caption{Success by ALFWorld task type with Qwen2.5-1.5B at update 160, evaluated on the same 140 tasks as Figure~\ref{fig:training-dynamics}. Axes scale linearly from 0\% at the center to 100\% at the boundary.}
    \label{fig:alfworld-task-types}
\end{figure}
\endgroup

\Needspace{4\baselineskip}
\subsection{Training dynamics}
\label{app:training-dynamics}

\begin{figure}[H]
    \centering
    \includegraphics[width=\textwidth]{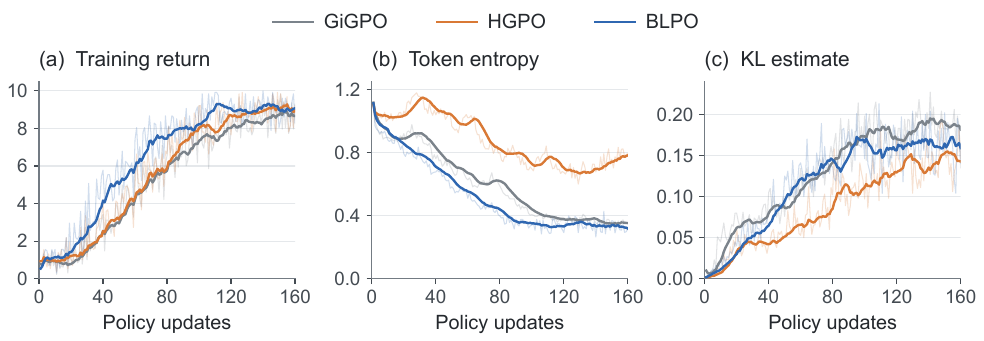}
    \caption{Training diagnostics on ALFWorld with Qwen2.5-1.5B, using one training run per method. Pale lines show per-update measurements; solid lines show 10-update trailing means.}
    \label{fig:training-diagnostics}
\end{figure}

Figure~\ref{fig:training-diagnostics} reports training diagnostics over all 160 policy updates. Return is the mean undiscounted environment reward of the sampled rollout batch, before the format penalty. Entropy is averaged over output tokens, rather than executable action categories. \method{} increases return more rapidly during early and middle training, while the gap to HGPO narrows toward the end. Its entropy decreases earlier and approaches GiGPO's level without a sustained decline in return. Meanwhile, the logged KL estimate relative to the reference policy levels off, with late values between those of GiGPO and HGPO.

\end{document}